\def\year{2020}\relax
\documentclass[letterpaper]{article} 
\usepackage{aaai20}  
\usepackage{times}  
\usepackage{helvet} 
\usepackage{courier}  
\usepackage[hyphens]{url}  
\usepackage{graphicx} 
\usepackage{graphicx}  
\usepackage{booktabs}
\usepackage{threeparttable}
\usepackage{multirow}
\usepackage{algorithm}
\usepackage{algorithmic}

\title{The SPPD System for Schema Guided Dialogue State Tracking Challenge}

\author{ \Large \textbf{Miao Li, Haoqi Xiong, Yunbo Cao}\\ 
Smart Platform Product Department, Tencent Inc, China \\
\{miaogli, haoqixiong, yunbocao\}@tencent.com 
}
 
\begin{document}

\maketitle

\begin{abstract}
This paper introduces one of our group's work on the Dialog System Technology Challenges 8 (DSTC8) \cite{DSTC8}, the SPPD system for Schema Guided dialogue state tracking challenge. This challenge, named as Track 4 in DSTC8, provides a brand new and  challenging dataset for developing scalable multi-domain dialogue state tracking algorithms for real world dialogue systems. We propose a zero-shot dialogue state tracking system for this task. The key components of the system is a number of BERT based zero-shot NLU models that can effectively capture semantic relations between natural language descriptions of services' schemas and utterances from dialogue turns. We also propose some strategies to make the system better to exploit information from longer dialogue history and to overcome the slot carryover problem for multi-domain dialogues. The experimental results show that the proposed system achieves a significant improvement compared with the baseline system.

\end{abstract}

\section{Introduction}

Task-oriented spoken dialogue systems (SDS) enable human users to acquire information or services through natural language conversations.
For complex tasks, such as finding restaurants or booking flights, the dialogue system usually needs to interact with a human user through multiple turns to find the user's goal and then provide a proper service.
The dialogue state tracking (DST) component has become a core component in modern dialogue systems.
It can provide a compact representation of a multi-turn conversation and help the dialogue policy to decide the next action to take.

Recently, the Dialogue State Tracking Challenges (DSTC) \cite{williams2013dialog,henderson2014second,henderson2014third,kim2017fourth,kim2016fifth} provide shared benchmarks to the research community and a variety of state tracking models are proposed.
There are rule-based models \cite{wang2013simple}, statistical models such as generative models \cite{thomson2010bayesian} and discriminative models \cite{lee2013structured}. 

The state of the art models are deep learning-based models. 
\cite{mrkvsic2016neural} proposed a DST model directly predict dialogue states based on natural language utterances instead of NLU results. They use neural networks to learn representations for user and system utterances as well as slot and values in the domain ontology. They do not need a delexicalisation vocabulary but they need to enumerate all entries in the ontology to detect the possible slot values mentioned in user utterances, which makes it not scalable to domains with large or unbounded possible values. 
\cite{rastogi2017scalable} proposed a model that chooses values from a much smaller candidate set rather than the whole ontology, but it needs an additional language understanding module to extract the candidate values.
Recently, end-to-end models are proposed to generate dialogue states through a seq2seq framework \cite{xu2018end,lei2018sequicity,wu2019transferable}. Through copy or pointer mechanism \cite{see2017get}, these models can effectively handle OOV values which may occur frequently in dynamic ontology for real work applications.

\section{Task Description}
Besides the large and dynamic ontology problems,
the real world dialogue systems, such as Google Assistant, Alexa and Siri, also need to support a large number of domains and even larger numbers of services.
There are an ever-increasing number of new domains or services created by third-party developers. 
Therefore, state tracking models that are scalable across multiple domains, and can easily generalize to new domains and services have attracted many research interests.
Track4 of DSTC8 provides a proper dataset for developing such models.

\subsection{Task Data Set}

The track4 dataset, named as Schema Guided Dialogue (SGD) dataset, has over 16000 dialogues in the training set covering 26 services which belong to 16 domains \cite{rastogi2019towards}.
The dataset defines a schema for each service. A schema defines names and natural language descriptions for intents and slots of its corresponding service.
A service may have more than one intents, such as ``Find a restaurant'' and ``Book a table" for the restaurant domain. The schema also defines the required slots and optional slots for each intent.
There are two type of slots in the schema, which are categorical slots and non-categorical slots.
For categorical slots, such as gender or day of the week, the schema provides all possible values that can be assigned to those slots.
For non-categorical slots, such as address or city, the schema nearly provides nothing about the possible values.

The ``state'' for a dialogue turn in this task is defined with 3 fields:
\begin{itemize}
\item ``Active Intent": this field represents a user's intent, and the candidate values for the intent field are provided by the schema. Accuracy is used to evaluate the proposed model for this field.
\item ``Requested Slots": this field indicates the slots requested by a user. A user may request more than one slot in a single turn. Macro-averaged F1 score is used to evaluate the proposed model for this field.
\item ``Slot Values": this field records the user's goal which is a set of constrains of slot value pairs provided by the user during the conversation. Average goal accuracy and Joint goal accuracy are used to evaluate the proposed model for this field.

\end{itemize}

The dialogues of the dataset are created by machine-machine interactions proposed by \cite{shah2018building}. 
The dataset contains both single domain dialogues and multi-domain dialogues. 
In a multi-domain dialogue, the user and the system will have multiple interactions talking about more than one domains, and information will be shared between these domains.
For each user turn of a dialogue, the proposed dialogue state tracking model must make predictions for all three fields, and metrics mentioned above are used to evaluate the model predictions.

The dataset provides rich yet different grained annotations. For each system turn in a dialogue, turn-level dialogue actions are provided. While for each user turn in a dialogue, only state-level annotations are provided.
Span information about non-categorical slots is provided both in user turn and system turn.

To test the model's ability for unseen services, the evaluation set contains many services, and consequently slots, which are not presented in the training set.

\subsection{Challenges of this task}

Based on our understanding, we discuss some challenges of this task here.

The first challenge is the zero-shot problem. The evaluation set may contain unseen domains or services, and the schemas of these unseen services are not accessible by the model during training.
The proposed model should be able to transfer knowledge from seen services in the training set to unseen services in the evaluation set only by the similarity of natural language descriptions.

The second challenge is the long-range dependency understanding problem.
Based on our observation of the dataset, about 10\% slot values in current dialogue state are mentioned beyond the current user utterance and the preceding system utterance.
So it is not enough to capture all the needed slot-values by a conventional turn-level-based natural language understanding (NLU) module. 

The third challenge is the slot carryover \cite{naik2018contextual} problem from the multi-domain dialogues. 
For example, when a user has booked a restaurant table for 2 people and plan to get there by a taxi, he would say ``Please find a taxi for me and my friend to get there''.
The system need to aware that the slot ``destination'' of the Taxi domain should be filled by the address of the restaurant just booked by the user,  and the value of slot ``number\_of\_riders'' of the Taxi domain should be equal to the value of slot ``number\_of\_seats'' from the Restaurant domain.

\section{Our Methods}

To overcome these challenges, we propose a system consists of an NLU module and a DST module. 
The NLU module is used to predict the active intent, requested slots and slot values according to the previous system turn and current user turn. 
And then, the DST module update the dialogue state based on the predictions from the NLU module.

We first introduce the basic implementation of the NLU module and the DST module which are designed to solve the zero-shot problem.
Then we propose some techniques to overcome the last two challenges.

\subsection{Basic NLU Module}

\begin{figure}[ht]
\centering
\includegraphics[scale=0.4]{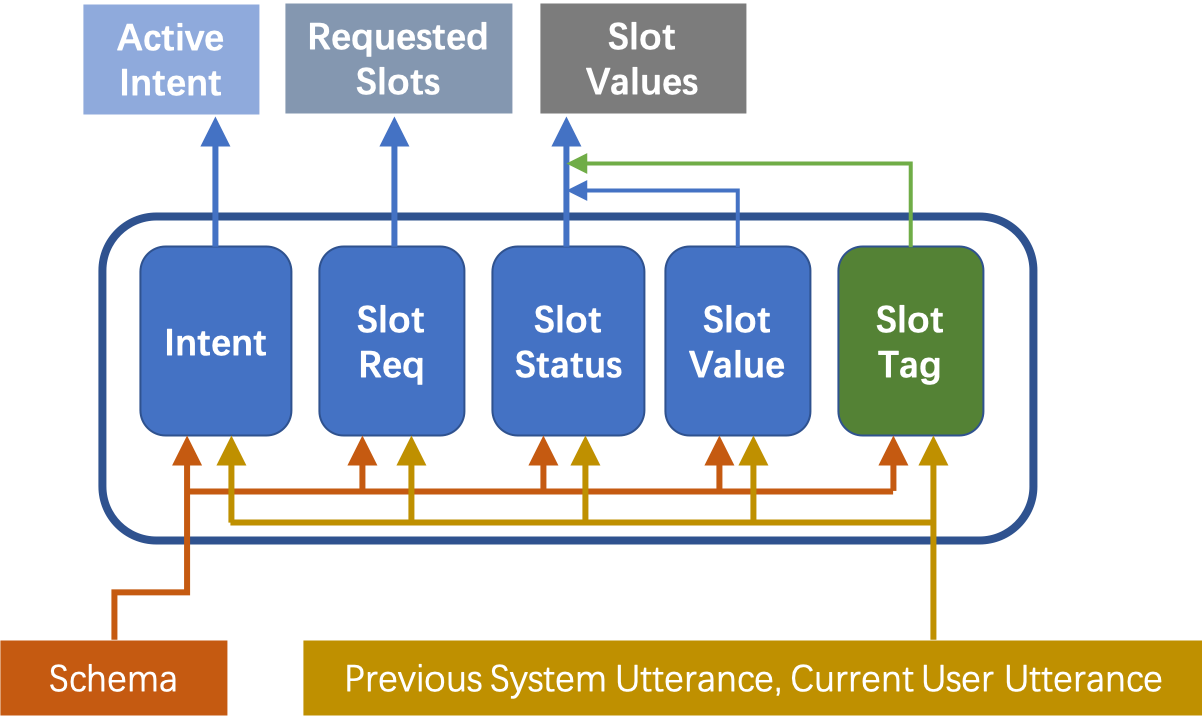}
\caption{The structure of the proposed NLU module}
\label{fig:nlu}
\end{figure}

The NLU module needs to predict all three fields, which are active intent, requested slots and the user goal which is represented as slot values for both categorical slots and non-categorical slots.
Here we propose a zero-shot NLU module consists of 5 models to generate NLU results.
The inputs of the NLU module are natural language descriptions of the service schema and dialogue contexts which consists of the previous system utterance and the current user utterance.
The structure of the NLU module is shown in Figure \ref{fig:nlu}.
We briefly introduce the function of each of the 5 models below.

1) Intent model: For a given service, we can collect all available intents from the service's schema and add a special ``NONE" for no intent to build a candidate set for current activate intent.
The intent model predicts a score for each candidate value and we choose the intent with highest score as current active intent.

2) Slot Request model: this model predicts a probability for each slot in the service schema. All slots with probability $>$ 0.5 are predicted as requested slots during inference.

3) Slot Status model: for each slot in the schema, this model predicts a distribution of size 3 denoting the probability of slot status. There are 3 status for a slot, which are ``None", ``Active" and ``Dontcare". 
The ``None" status means that the value of the slot should stay the same.
The ``Dontcare" status means that the slot should take the special value ``dontcare".
The ``Active" status means that we should assign a value for this slot. The following two models will extract a value for the active slot.

4) Slot Value model: this model is used to extract values for categorical slots. For a categorical slot, the model predicts a probability for each of its possible values defined in the schema . The value with the largest probability is considered as the predicted value for the slot.

5) Slot Tagging model: this model predicts the a span for each non-categorical slots in the schema.

For the sake of simplicity, we will summarize the last two models as ``value extraction models" in the following of this paper. 

We use BERT \cite{devlin2018bert} as our base model to build the 5 models in the NLU module. BERT is a multi-layer bidirectional Transformer network pretrained on very large corpus.
With the self-attention connections, BERT can capture bidirectional long-range dependencies for natural language.
The fine-tuned BERT models have achieved state of the art performances on many NLP tasks, such as sentence (single sentence or sentence pair) classification tasks in GLUE \cite{wang2018glue} and machine reading comprehension tasks in SQUAD \cite{rajpurkar2016squad,rajpurkar2018know}.

The input to BERT is a sequence of tokens prepended by a special token [CLS].
The final hidden state of token [CLS] is used as the aggregate representation of the input sequence, while the final hidden states of other tokens are used as token-level representations.
To differentiate tokens from the pair of sentences, the input layer of BERT adds an additional segment embedding.

The 1-4 models in the NLU module are classification based models, the Slot Status model is a 3-class model while the other 3 models are 2-class models.
We borrow the idea of sentence pair classification BERT model to build the first 4 classification based models in the NLU module.


Unlike \cite{chao2019bert} or \cite{rastogi2019towards}, using preceding system utterance and current user utterance as the input sentence pair to BERT model,
we use natural language descriptions of the slots/intents and the dialogue utterances as the input sentence pair.
We believe that using a single BERT model to encode the schema descriptions and the dialogue utterances would better capture the semantic relations between the schema and the dialogue utterance.

For the 5th model, the slot tagging model, we treat the description of a non-categorical slot as the question and the user utterance as the document, then we use the exact same structure of BERT model for SQUAD to extract spans for non-categorical slots. 
If a non-categorical slot is not mentioned in the user utterance, then the [CLS] token will have the biggest probabilities for both start and end logits for the span. 
This model can also produce n-best results with probabilities.

The input sentence pairs for different models are listed in Table \ref{tab_model_input}. 
The first part of input sequence contains information from the schema, and the second part contains information from dialogue context.
For slot request model and slot tagging model, we only use current user utterance as the dialogue context.
We think current user utterance contains enough information for detecting the requested slots.
Using only current user utterance may miss the values of non-categorical slots mentioned in preceding system utterances, we propose a value retrieval strategy which will be talked later  to solve this problem.
For the other models, we use both preceding system utterance and current user utterance to build the model inputs.

\begin{table}[!htbp]
\centering
\caption{The input sequences of different NLU models}
\begin{threeparttable}
\begin{tabular}{ccc}
\toprule
Model & Sequence 1 & Sequence 2 \\
\midrule
Intent & Intent name + description & PSU + CUU\tnote{*} \\
Slot request & Slot name + description & CUU \\
Slot status & Slot name + description & PSU + CUU \\
Slot value & Slot name + value name & PSU + CUU \\
Slot tagging & Slot name + description & CUU \\
\bottomrule
\end{tabular}

\begin{tablenotes}
\footnotesize
\item[*] PSU represents Preceding System Utterance, and CUU represents Current User Utterance
\end{tablenotes}
\end{threeparttable}

\label{tab_model_input}
\end{table}

\subsection{Basic DST Module}

The basic DST module is a simple rule-based module. It uses the output of the NLU module to update the dialogue state from the previous turn.
The DST module will always update the active intent and requested slots predicted by the NLU module.
For user goal field, If the NLU module predicts a slot and its value, whether be ``dontcare'' or extracted by the value extraction models, the slot and value pair will be used to update the dialogue state.
Otherwise, the user goal of the slot stays the same as the previous turn.

\subsection{Improvement Strategies}

Based on our system structure, we propose some effective strategies to further improve the performance of our systems.

\subsubsection{Previous Intent Dropout}

Actually, using only preceding system utterance and current user utterance as dialogue context to do the NLU prediction is not enough for some models, especially for the Intent model.
For example, there are two different intents, which are ``SearchOnewayFlight'' and ``SearchRoundtripFlights'', in the ``Flight'' domain.
But these two intents both have a lot of slots, such as  ``origin city'', ``destination city'' and ``departure time'', in common.
When we only looking at a preceding system utterance ``Which city are you going to?'' and a user response ``I would like to go to New York City'', we can hardly distinguish these two intents.

Instead of adding longer dialogue turn context to the model, we add the intent of previous state to the model to solve this problem.
But simply adding the previous intent to the model will bring the mismatch between training and inference,
since we use gold previous intent during training, while using model predicted previous intent during inference.
Here we propose a special dropout method which is called ``previous intent dropout'' to mitigate the mismatch.
The ``previous intent dropout'' is only used in training phase, we simply dropout the previous intent (do not add it to the input sequence) follow a Bernoulli distribution.
We will show that this simple yet method is effective for intent prediction later in the experiment part.

\subsubsection{Combine the Outputs from User Goal Models in a Probabilistic Way}

In our proposed NLU module, we need to combine the outputs of three models (slot status model, slot value model, slot tagging model) to generate the full slot value constrains for the user goal.
A straightforward way is combine these models in a pipeline way \cite{chao2019bert,rastogi2019towards}, the slot status model can be treated as the first stage, while the value extraction models (slot value model and the slot tagging model) can be treated as the second stage.
When the slot status model predicts an ``Active'' status for a slot in the first stage, we go to the second stage and extract the value for the slot by value extraction models.

In the two-stage situation, the first stage decision is made based solely on the slot status model.
This may cause some problems when the two stage models have some disagreements. 
For example, the slot status model may think one slot is in an active status while the value extraction model predicts a very low probability of its value, or vice versa the value extraction model predicts a very high probability of a value but the slot status model thinks the slot is non-active.

We believe combining the information from these two stage models may help us make a better decision.
So we propose a probabilistic method to combine the outputs of the two stage models.
In the probabilistic method, we first make decision based on the slot status model. At second stage, we combine the slot status model and value extraction models to assign value for the slot. The proposed probabilistic algorithm is illustrated in Algorithm \ref{alg:ProbAvg}. 
We only consider the max probability value $V_m$ and its probability $P_m$ extracted by the value extraction models (slot value model or slot tagging model) in this algorithm.
At the second stage, if the average probability of $P_{Active}$ and $P_m$ is larger than a threshold (which is set to 0.5), the value $V_m$ will be assigned to slot $S$ even if the slot status model predict a ``None'' status. If the slot status model predict an ``Active'' status but the average probability of $P_{Active}$ and $P_m$ is less than the threshold, we will assign a special value ``[]'' to the slot, which indicates the slot is active but without explicit value.

\begin{algorithm}[htb] 
\caption{Probabilistic Average algorithm to combine the model outputs} 
\label{alg:ProbAvg} 
\begin{algorithmic}[1]
\REQUIRE ~~\\
\STATE Target slot $S$;\\
\STATE The probability distribution produced by slot status model: \{ $P_{None}, P_{Dontcare}, P_{Active}$ \};\\
\STATE Max probability value and its probability extracted by value extraction models: $V_m$, $P_m$,;\\
\STATE A Threshold $T$
\ENSURE ~~\\ 
The value of target slot $S$
\IF{$ max(P_{None}, P_{Dontcare}, P_{Active}) == P_{None}$}
  \IF{$(P_{Active} + P_m)/2 > T$}
    \RETURN $S=V_m$;
  \ELSE
    \RETURN $S=None$;
  \ENDIF
\ELSIF {$ max(P_{None}, P_{Dontcare}, P_{Active}) == P_{Active}$}
  \IF{$(P_{Active} + P_m)/2 > T$}
    \RETURN $S=V_m$;
  \ELSE
    \RETURN $S=[]$;
  \ENDIF
\ELSE
  \RETURN $S=``Dontcare"$;
\ENDIF

\end{algorithmic}
\end{algorithm}

Then, we can produce a frame structure as the NLU results which will be used to update the dialogue state by the DST module.
It's worth to notice that each slot has three available status (a little different with the status in slot status model) to choose in this frame structure.
The first status is none-active ($S=None$), which means the slot will not be updated in the state.
The second status is active with explicit value ($S=``Dontcare''$ or $S=V_m$), and slot value pair will be used to update the dialogue state.
The third status is active without explicit value ($S=[]$), and the value need to be retrieved through longer dialogue history. 
We will also talk this value retrieval strategy in the following part of this sub-section.

\subsubsection{Value Retrieval}

Not only the intent understanding mentioned above needs longer dialogue history, the user goal understanding also needs to exploit longer dialogue history, especially in this task.
Based on the slot span labels of non-categorical slots, we can know exactly which part of the dialogue context does a non-categorical slot in the user goal come from.
For example, some non-categorical slots are mentioned in current user utterance, some are mentioned in preceding system utterance.
There even exist some non-categorical slots that are mentioned before preceding system utterances, or come from other services in earlier dialogue contexts.
For non-categorical slots in in the dialogue state, We find that there are only about 40\% of them mentioned in the current user utterance, about 50\% of them mentioned in preceding system utterance, and the last 10\% of them come from even longer dialogue history.

We propose a simple yet effective value retrieval strategy for DST module to overcome this long-range dependency understanding problem. 
As we mentioned before, the slot may have a special status, which is ``active without explicit value'', in the outputs of our NLU module.
A slot in this status means that we should update slot's value in the dialogue state but the value is not mentioned in current user utterance.
When the DST module finds a slot is in ``active without explicit value'' status, the DST module will search the entire dialogue history before current user utterance in a reversed order to find if a value for this slot is mentioned in a previous system action.
If a value is found, the DST module will assign the value to this slot.

\subsubsection{Slot Carryover}

According to our statistics of the dataset, there are about 6\% non-categorical slots in the dialogue state are actually mentioned in a different service in previous dialogue history. We call this a slot carryover problem followed by \cite{naik2018contextual}.
To solve the slot carryover problem, we need to solve two sub-problems.
The first sub-problem is to generate a candidate carryover slot set for a target slot. Each slot in the candidate set exists a carryover relation with the target slot.
The second sub-problem is to make the carryover decision, whether a value of a slot in the candidate slot set should carryover to the target slot based on the conversation context.

We cast the first sub-problem as a 2-class classification problem, and train a carryover relation extraction model to solve it.
The carryover relation extraction model should be able to predict whether there exist a carryover relation between two given slots based on their names and descriptions.
We assume the carryover relation is a symmetrical and transitive binary relation of two slots, and build the training data from the original training set based on this assumption.
Then we train a BERT-based classification model on the training data to determine whether two slots exists a carryover relation.
The concatenation of two slots' name and description are used as input sequence to the model.
In the inference phase, we first use this carryover relation extraction model to generate a candidate slot set for every slot in the schema of the evaluation set.

For the second sub-problem, we follow the same idea proposed in the value retrieval strategy.
If a slot is in ``active without explicit value'' status but the value is not found by the value retrieval strategy,
we try to search values of slots in the target slot's candidate set from previous system actions. If a value of a slot in the target slot's candidate carryover slot set is found, the value will be assigned to the target slot by the DST module.

\section{Experiments}

\subsection{Datasets}

We evaluated our proposed zero-shot state tracking system on the Schema-Guided Dialogue dataset of the DSTC8 track4.
There are single domain dialogues and multi-domain dialogues in both training set and dev set. 
We use all the dialogues in the training set for model training and report our results on the whole dev set.

To train our proposed NLU models, we automatically generate NLU label based on the state-level annotations from the training set.
For intent and requested slots fields, the labels of current dialogue state are set as the labels for current turn.
For slot values in user goal field, we use the difference between the user goal for the current turn and preceding user turn as the current turn's NLU label.

\subsection{Implementation details}

We use the BERT-Base-uncased pre-trained model as our base model and all the input text are lower-cased before feeding to the model.
We found using dialogue actions instead of original utterances for system turns leads a better performance, so dialogue actions for system turn are used in all the systems reported below. 
The 5 NLU models were trained independently using automatically generated NLU training data from entire training set.
The ADAM optimizer was used to update all layers in the models. The initial learning rate was set to 2e-5 and linear learning rate decay was used. 
All models are trained for 3 epochs.

Various rates of previous intent dropout are experimented and reported in following section.

\begin{table*}[h]
\centering
\caption{The overall performance of our system}
\begin{threeparttable}
\begin{tabular}{lcccc}
\toprule
\multirow{2}*{Model} & Intent Accuracy & Average Goal Accuracy & Joint Goal Accuracy & Requested Slot F1  \\
  & All(Seen/Unseen)  & All(Seen/Unseen)  & All(Seen/Unseen)  & All(Seen/Unseen)  \\
\midrule
Baseline \cite{rastogi2019towards} & 90.8(NA/NA) & 74.0(NA/NA) & 41.1(NA/NA) & 97.3(NA/NA) \\
Baseline* & 91.8(96.6/85.5) & 72.1(80.2/61.4) & 43.8(54.9/24.3) & 96.6(99.5/92.8) \\
Our Basic System & 98.5(98.4/98.6) & 75.5(75.2/75.8) & 41.9(41.9/41.9) & 99.0(99.7/98.0) \\
\hspace{1em}+VR PSU & 98.5(98.4/98.6) & 86.3(86.4/86.2) & 65.2(67.2/62.6) & 99.0(99.7/98.0) \\
\hspace{2em}+VR LS & 98.5(98.4/98.6) & 88.4(88.5/88.3) & 69.6(71.9/66.7) & 99.0(99.7/98.0) \\
\hspace{3em}+SCO & 98.5(98.4/98.6) & 95.8(96.5/94.9) & 82.1(85.9/77.2) & 99.0(99.7/98.0) \\
\hspace{4em}+ProbAvg & 98.5(98.4/98.6) & 96.4(97.4/95.2) & 84.0(88.0/78.9) & 99.0(99.7/98.0) \\
BERT Large + All & 98.2(98.4/97.8) & 96.4(97.8/94.7) & 88.1(89.9/85.6) & 99.3(99.8/98.6) \\
Ensemble + All & 98.8(98.9/98.6) & 96.6(97.4/95.5) & 90.0(91.6/87.8) & 99.5(99.8/99.0) \\

\bottomrule
\end{tabular}

\begin{tablenotes}
\footnotesize
\item +VR PSU means adding value retrieval strategy for preceding system utterance. \\
+VR LS means adding value retrieval strategy for longer history. \\
+SCO means adding slot carryover strategy \\
+ProbAvg means adding probabilistic combination strategy \\
+All means adding all proposed strategies \\
\end{tablenotes}
\end{threeparttable}

\label{results}
\end{table*}

\subsection{Results}

The experimental results are reported here. First, we evaluate the proposed previous intent dropout strategy for intent prediction. Then we report the overall performances for our system enhanced with various improvement strategies.

\subsubsection{Performance on intent classification}

\begin{figure}[ht]
\centering
\includegraphics[scale=0.5]{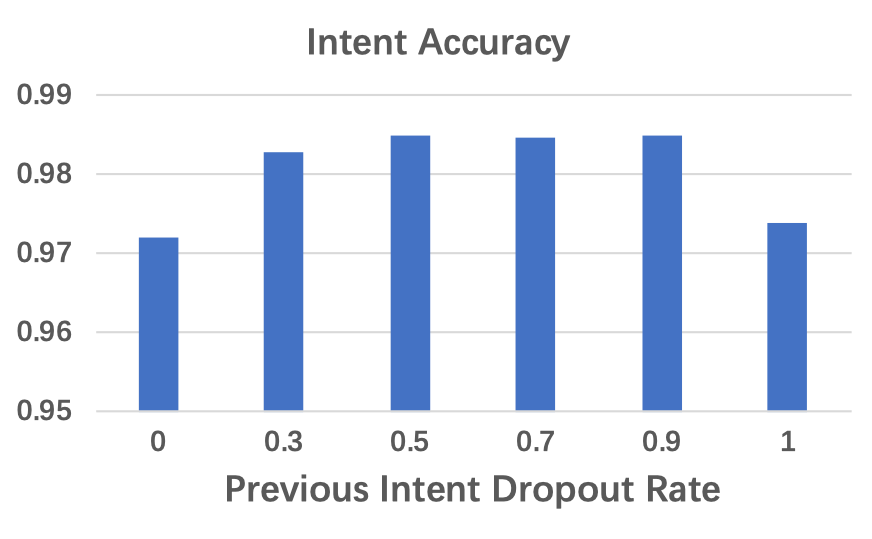}
\caption{Intent accuracy for different previous intent dropout rate}
\label{fig:dpi}
\end{figure}

First we report the experimental results of our proposed previous intent dropout strategy. 
Figure \ref{fig:dpi} shows the performance of different intent models trained by different previous intent dropout rate.
The dropout rate 1.0 means we never add previous intent information to the model in neither training phase nor evaluation phase.
For dropout rate smaller than 1.0, we dropout the previous intent during training and always add previous intent predicted by the model during evaluation.

If we never add previous intent to the model, we may lack enough information to make the prediction. 
But if we always add previous intent to the model, the learned model may make the prediction largely based on the previous intent information, and suffers from error propagation during inference.
In our experiments, adding a proper previous intent dropout can always improve the intent accuracy compared with baseline models.
A considerable large dropout rate (0.9) achieves the best performance, which is out of our expectation.
The larger dropout rate may make the learned model pay more attention to current dialogue context thus more robust to error propagation.

Applied with this strategy, our system achieves the best intent accuracy in the evaluation set.

\subsubsection{Performance on the overall task}

Table \ref{results} shows the overall performance of the baseline system \cite{rastogi2019towards} and our systems.
To show the zero-shot ability for the compared systems, the performance about the services that are seen/unseen in the training data are also provided.
Since the performance about the seen/unseen services are not provided by the original paper of the baseline system, we list the performance of our reimplementation here as ``baseline*''.
Without any improvement strategies, which means the system can only catch non-categorical slots that mentioned in current user utterances, our basic system still surpasses the baseline model, which is able to extract values for non-categorical slots that are mentioned both in current user utterances and preceding system utterances. We think this is because that using a single BERT model to encode the schema and the dialogue utterances can better capture the semantic relations through the attention mechanism.
It is worth noting that the performance gap between the seen services and the unseen services of our system is much smaller than the baseline system. Which further demonstrates the strong zero-shot ability of our system. 

By adding the value retrieval strategy and slot carryover strategy, the proposed system can access information from longer dialogue history and other services, thus improving both the average and joint goal accuracies. The proposed probabilistic combination strategy also allow the system to make better decisions by combining the outputs from different models which leads to an improvement both in average and joint goal accuracies. Using BERT-Large as the pretrained model can further improve the overall performances. Finally we build an ensemble model by averaging various models finetuned on various pretrained models such as BERT-Base, BERT-Large, and Roberta-Large.

\section{Conclusion}

In this paper, we introduce our DST system for a challenging zero-shot schema guided dialogue state tracking task.
Our NLU module uses BERT to encode the schema descriptions and dialogue utterances together to better capture the semantic relations between them, which makes the module easily transferred to unseen domains and services. Besides, we propose some strategies to combine the outputs of our NLU models in a probabilistic way, to make better use of long dialogue history, and to solve the slot carryover problem. Enhanced with these strategies, our proposed zero-shot dialogue state tracking system achieves significant improvement compared with the baseline system.

\bibliography{ref}

\begin{thebibliography}{}

\bibitem[\protect\citeauthoryear{Chao and Lane}{2019}]{chao2019bert}
Chao, G.-L., and Lane, I.
\newblock 2019.
\newblock Bert-dst: Scalable end-to-end dialogue state tracking with
  bidirectional encoder representations from transformer.
\newblock {\em arXiv preprint arXiv:1907.03040}.

\bibitem[\protect\citeauthoryear{Devlin \bgroup et al\mbox.\egroup
  }{2018}]{devlin2018bert}
Devlin, J.; Chang, M.-W.; Lee, K.; and Toutanova, K.
\newblock 2018.
\newblock Bert: Pre-training of deep bidirectional transformers for language
  understanding.
\newblock {\em arXiv preprint arXiv:1810.04805}.

\bibitem[\protect\citeauthoryear{Henderson, Thomson, and
  Williams}{2014a}]{henderson2014second}
Henderson, M.; Thomson, B.; and Williams, J.~D.
\newblock 2014a.
\newblock The second dialog state tracking challenge.
\newblock In {\em Proceedings of the 15th Annual Meeting of the Special
  Interest Group on Discourse and Dialogue (SIGDIAL)},  263--272.

\bibitem[\protect\citeauthoryear{Henderson, Thomson, and
  Williams}{2014b}]{henderson2014third}
Henderson, M.; Thomson, B.; and Williams, J.~D.
\newblock 2014b.
\newblock The third dialog state tracking challenge.
\newblock In {\em 2014 IEEE Spoken Language Technology Workshop (SLT)},
  324--329.
\newblock IEEE.

\bibitem[\protect\citeauthoryear{Kim \bgroup et al\mbox.\egroup
  }{2016}]{kim2016fifth}
Kim, S.; D'Haro, L.~F.; Banchs, R.~E.; Williams, J.~D.; Henderson, M.; and
  Yoshino, K.
\newblock 2016.
\newblock The fifth dialog state tracking challenge.
\newblock In {\em 2016 IEEE Spoken Language Technology Workshop (SLT)},
  511--517.
\newblock IEEE.

\bibitem[\protect\citeauthoryear{Kim \bgroup et al\mbox.\egroup
  }{2017}]{kim2017fourth}
Kim, S.; D’Haro, L.~F.; Banchs, R.~E.; Williams, J.~D.; and Henderson, M.
\newblock 2017.
\newblock The fourth dialog state tracking challenge.
\newblock In {\em Dialogues with Social Robots}. Springer.
\newblock  435--449.

\bibitem[\protect\citeauthoryear{Kim \bgroup et al\mbox.\egroup }{2019}]{DSTC8}
Kim, S.; Galley, M.; Gunasekara, C.; and Lee, S.
\newblock 2019.
\newblock The eighth dialog system technology challenge.
\newblock {\em arXiv preprint}.

\bibitem[\protect\citeauthoryear{Lee}{2013}]{lee2013structured}
Lee, S.
\newblock 2013.
\newblock Structured discriminative model for dialog state tracking.
\newblock In {\em Proceedings of the SIGDIAL 2013 Conference},  442--451.

\bibitem[\protect\citeauthoryear{Lei \bgroup et al\mbox.\egroup
  }{2018}]{lei2018sequicity}
Lei, W.; Jin, X.; Kan, M.-Y.; Ren, Z.; He, X.; and Yin, D.
\newblock 2018.
\newblock Sequicity: Simplifying task-oriented dialogue systems with single
  sequence-to-sequence architectures.
\newblock In {\em Proceedings of the 56th Annual Meeting of the Association for
  Computational Linguistics (Volume 1: Long Papers)},  1437--1447.

\bibitem[\protect\citeauthoryear{Mrk{\v{s}}i{\'c} \bgroup et al\mbox.\egroup
  }{2016}]{mrkvsic2016neural}
Mrk{\v{s}}i{\'c}, N.; S{\'e}aghdha, D.~O.; Wen, T.-H.; Thomson, B.; and Young,
  S.
\newblock 2016.
\newblock Neural belief tracker: Data-driven dialogue state tracking.
\newblock {\em arXiv preprint arXiv:1606.03777}.

\bibitem[\protect\citeauthoryear{Naik \bgroup et al\mbox.\egroup
  }{2018}]{naik2018contextual}
Naik, C.; Gupta, A.; Ge, H.; Mathias, L.; and Sarikaya, R.
\newblock 2018.
\newblock Contextual slot carryover for disparate schemas.
\newblock {\em arXiv preprint arXiv:1806.01773}.

\bibitem[\protect\citeauthoryear{Rajpurkar \bgroup et al\mbox.\egroup
  }{2016}]{rajpurkar2016squad}
Rajpurkar, P.; Zhang, J.; Lopyrev, K.; and Liang, P.
\newblock 2016.
\newblock Squad: 100,000+ questions for machine comprehension of text.
\newblock {\em arXiv preprint arXiv:1606.05250}.

\bibitem[\protect\citeauthoryear{Rajpurkar, Jia, and
  Liang}{2018}]{rajpurkar2018know}
Rajpurkar, P.; Jia, R.; and Liang, P.
\newblock 2018.
\newblock Know what you don't know: Unanswerable questions for squad.
\newblock {\em arXiv preprint arXiv:1806.03822}.

\bibitem[\protect\citeauthoryear{Rastogi \bgroup et al\mbox.\egroup
  }{2019}]{rastogi2019towards}
Rastogi, A.; Zang, X.; Sunkara, S.; Gupta, R.; and Khaitan, P.
\newblock 2019.
\newblock Towards scalable multi-domain conversational agents: The
  schema-guided dialogue dataset.
\newblock {\em arXiv preprint arXiv:1909.05855}.

\bibitem[\protect\citeauthoryear{Rastogi, Hakkani-T{\"u}r, and
  Heck}{2017}]{rastogi2017scalable}
Rastogi, A.; Hakkani-T{\"u}r, D.; and Heck, L.
\newblock 2017.
\newblock Scalable multi-domain dialogue state tracking.
\newblock In {\em 2017 IEEE Automatic Speech Recognition and Understanding
  Workshop (ASRU)},  561--568.
\newblock IEEE.

\bibitem[\protect\citeauthoryear{See, Liu, and Manning}{2017}]{see2017get}
See, A.; Liu, P.~J.; and Manning, C.~D.
\newblock 2017.
\newblock Get to the point: Summarization with pointer-generator networks.
\newblock {\em arXiv preprint arXiv:1704.04368}.

\bibitem[\protect\citeauthoryear{Shah \bgroup et al\mbox.\egroup
  }{2018}]{shah2018building}
Shah, P.; Hakkani-T{\"u}r, D.; T{\"u}r, G.; Rastogi, A.; Bapna, A.; Nayak, N.;
  and Heck, L.
\newblock 2018.
\newblock Building a conversational agent overnight with dialogue self-play.
\newblock {\em arXiv preprint arXiv:1801.04871}.

\bibitem[\protect\citeauthoryear{Thomson and Young}{2010}]{thomson2010bayesian}
Thomson, B., and Young, S.
\newblock 2010.
\newblock Bayesian update of dialogue state: A pomdp framework for spoken
  dialogue systems.
\newblock {\em Computer Speech \& Language} 24(4):562--588.

\bibitem[\protect\citeauthoryear{Wang and Lemon}{2013}]{wang2013simple}
Wang, Z., and Lemon, O.
\newblock 2013.
\newblock A simple and generic belief tracking mechanism for the dialog state
  tracking challenge: On the believability of observed information.
\newblock In {\em Proceedings of the SIGDIAL 2013 Conference},  423--432.

\bibitem[\protect\citeauthoryear{Wang \bgroup et al\mbox.\egroup
  }{2018}]{wang2018glue}
Wang, A.; Singh, A.; Michael, J.; Hill, F.; Levy, O.; and Bowman, S.~R.
\newblock 2018.
\newblock Glue: A multi-task benchmark and analysis platform for natural
  language understanding.
\newblock {\em arXiv preprint arXiv:1804.07461}.

\bibitem[\protect\citeauthoryear{Williams \bgroup et al\mbox.\egroup
  }{2013}]{williams2013dialog}
Williams, J.; Raux, A.; Ramachandran, D.; and Black, A.
\newblock 2013.
\newblock The dialog state tracking challenge.
\newblock In {\em Proceedings of the SIGDIAL 2013 Conference},  404--413.

\bibitem[\protect\citeauthoryear{Wu \bgroup et al\mbox.\egroup
  }{2019}]{wu2019transferable}
Wu, C.-S.; Madotto, A.; Hosseini-Asl, E.; Xiong, C.; Socher, R.; and Fung, P.
\newblock 2019.
\newblock Transferable multi-domain state generator for task-oriented dialogue
  systems.
\newblock {\em arXiv preprint arXiv:1905.08743}.

\bibitem[\protect\citeauthoryear{Xu and Hu}{2018}]{xu2018end}
Xu, P., and Hu, Q.
\newblock 2018.
\newblock An end-to-end approach for handling unknown slot values in dialogue
  state tracking.
\newblock {\em arXiv preprint arXiv:1805.01555}.

\end{thebibliography}
\bibliographystyle{aaai}
\end{document}